# Word, graph and manifold embedding from Markov processes

Tatsunori B. Hashimoto
MIT CSAIL
Cambridge, MA 02139
thashim@csail.mit.edu

David Alvarez-Melis
MIT CSAIL
Cambridge, MA 02139
davidam@csail.mit.edu

Tommi S. Jaakkola
MIT CSAIL
Cambridge, MA 02139
tommi@mit.edu

September 18, 2015

## Abstract

Continuous vector representations of words and objects appear to carry surprisingly rich semantic content. In this paper, we advance both the conceptual and theoretical understanding of word embeddings in three ways. First, we ground embeddings in semantic spaces studied in cognitive-psychometric literature and introduce new evaluation tasks. Second, in contrast to prior work, we take *metric recovery* as the key object of study, unify existing algorithms as consistent metric recovery methods based on co-occurrence counts from simple Markov random walks, and propose a new recovery algorithm. Third, we generalize metric recovery to graphs and manifolds, relating co-occurence counts on random walks in graphs and random processes on manifolds to the underlying metric to be recovered, thereby reconciling manifold estimation and embedding algorithms. We compare embedding algorithms across a range of tasks, from nonlinear dimensionality reduction to three semantic language tasks, including analogies, sequence completion, and classification.

## 1  Introduction

Continuous vector representations of words, objects, and signals have been widely adopted across areas, from natural language processing and computer vision to speech recognition. Methods for estimating these representations such as neural word embeddings [3, 14, 12] are typically simple and scalable enough to be run on large corpora, yet result in word vectors that appear to capture syntactically and semantically meaningful properties. Indeed, analogy tasks have become *de facto* benchmarks to assess the semantic richness of word embeddings [8, 12]. However, theoretical understanding of why they work has lagged behind otherwise intriguing empirical successes.

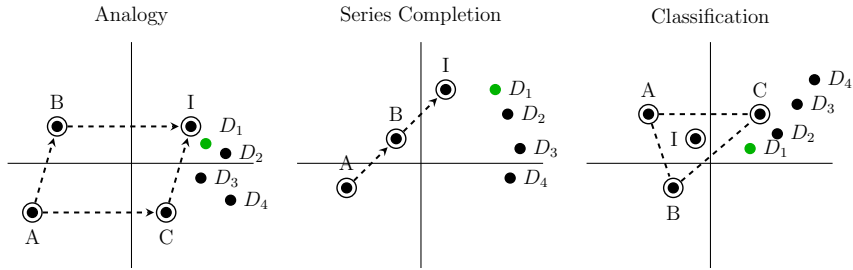

Figure 1: Sternberg's model for inductive reasoning with embeddings (A, B, C are given, I is the ideal point and D are the choices. The correct answer is shaded green).

Several recent contributions have aimed at bringing a better understanding of word embeddings, their properties, and associated algorithms [8, 5, 9, 1]. For example, [9] showed that the global minimum of the skip-gram method with negative sampling [12] implicitly factorizes a shifted version of the point-wise mutual information (PMI) matrix of word-context pairs. Arora et al. [1] explored links between random walks and word embeddings, relating them to contextual (probability ratio) analogies, under specific (isotropic) assumptions about word vectors.

In this paper, we extend the conceptual and theoretical understanding of word embeddings in three ways. First, we ground word embeddings to semantic spaces studied in cognitive-psychometric literature and consider three inductive reasoning tasks for evaluating the semantic content in word vectors, including analogies (previously studied) but also two new tasks, sequence completion and classification. We demonstrate existing and proposed algorithms perform well across these tasks. Second, in contrast to [1], we take *metric recovery* as the key object of study and unify existing algorithms as consistent metric recovery methods based on log(co-occurrence) counts arising from simple Markov random walks. Motivated by metric recovery, we also introduce and demonstrate a direct regression method for estimating word embeddings. Third, we generalize metric recovery to graphs and manifolds, directly relating co-occurence counts for random walks on graphs and random processes on manifolds to the underlying metric to be recovered.

## 2 Word vectors and semantic spaces

Semantic spaces, i.e., vector spaces where semantically related words are close to each other, have long been an object of study in the psychometrics and cognitive science communities [19, 21]. Rumelhart and Abrahamson proposed that vector word representations derived from semantic similarity along with vector addition could predict response choices in analogy questions [19]. This hypothesis was verified in three ways: by solving analogies using embeddings derived from survey data; observations that human mistake rates followed an exponential decay in embedded distance from the true solution; and ability of study subjects to answer analogies consistent with an embedding for nonexistent animals after training [19].

Sternberg further proposed that general inductive reasoning was based upon *metric embeddings* and tested two additional language tasks, series completion and classification (see Figure 1) [21]. In the completion task, the goal is to predict which word should come next in a series of words (e.g. given *penny*, *nickel*, *dime*, the answer should be *quarter*). In the classification task, the goal is to choose the word that best fits a set of given words. For example, given *zebra, giraffe* and *goat*, and candidate choices *dog, mouse, cat* and *deer*, the answer would be *deer* since it fits the first three terms best. Sternberg proposed that, given word embeddings, a subject solves the series completion problem by finding the next point in the line defined by the given words, and solves the classification task by finding the candidate word closest to the centroid of the given words. A reproduction of Sternberg's original graphical depiction of the three induction tasks is given in Figure 1. As with analogies, we find that word embedding methods perform surprisingly well at these additional tasks. For example, in the series completion task, given "body, arm, hand" we find the completion to be "fingers".

Many of the embedding algorithms are motivated by the *distributional assumption* [6]: words appearing in similar contexts across a large corpus should have similar vector representations. Going beyond this hypothesis, we follow the psychometric literature more closely and take metric recovery as the key object of study, unifying and extending embedding algorithms from this perspective.

## 3 Recovering semantic distances with word embedding

We begin with a simple model proposed in the literature [2] where word co-occurences over adjacent words represent semantic similarity and generalize the model in later sections. Our corpus consists of $m$ total words across $s$ sentences over a $n$ word vocabulary where each word is given a coordinate in a latent word vector space $\{x_1, \ldots, x_n\} \in \mathbb{R}^d$. For each sentence $s$ we consider a Markov random walk, $X_1, \ldots, X_{m_s}$, with the following transition function

$$\mathbb{P}(X_t = x_j | X_{t-1} = x_i) = \frac{\exp(-||x_i - x_j||_2^2/\sigma^2)}{\sum_{k=1}^{n} \exp(-||x_i - x_k||_2^2/\sigma^2)}. \tag{1}$$

## 3.1 Log co-ocurrence as a metric

Suppose we observe the Gaussian random walk (Eq. 1) over a corpus with $m$ total words and define $C_{ij}$ as the number of times for which $X_t = x_j$ and $X_{t-1} = x_i$.[1] By the Markov chain law of large numbers, as $m \to \infty$,

$$-\log\left(C_{ij}/\sum_{k=1}^{n} C_{ik}\right) \xrightarrow{p} ||x_i - x_j||_2^2/\sigma^2 + \log(Z_i)$$

where $Z_i = \sum_{k=1}^{n} \exp(-||x_i - x_k||_2^2/\sigma^2)$ (See Supplementary Lemma S1.1).

More generally, consider the following limit that relates log co-occurence to word embeddings

**Lemma 1.** *Let $C_{ij}$ be a co-occurence matrix over a corpus of size $m$ and $x$ be coordinates of words in the latent semantic space, then there exists a sequence $a_i^m$ and $b_j^m$ such that as $m \to \infty$,*

$$-\log(C_{ij}) - a_i^m \xrightarrow{p} ||x_i - x_j||_2^2 + b_j^m.$$

The Gaussian random walk above is a special case of this limit; we will show that random walks on graphs and some topic models fulfill this metric recovery limit.

## 3.2 Consistency of word embeddings

Appying this to three word embedding algorithms, we show the conditions of Lemma 1 are sufficient to ensure that the true embedding $x$ is a global minimum.
**GloVe:** The Global Vectors (`GloVe`) [17] method for word embedding optimizes the objective function

$$\min_{\widehat{x},\widehat{c},a,b} \sum_{i,j} f(C_{ij})(2\langle \widehat{x}_i, \widehat{c}_j \rangle + a_i + b_j - \log(C_{ij}))^2$$

with $f(C_{ij}) = \min(C_{ij}, 10)^{3/4}$. If we rewrite the bias terms as $a_i = \widehat{a}_i - ||\widehat{x}_i||_2^2$ and $b_j = \widehat{b}_j - ||\widehat{c}_j||_2^2$, we obtain the equivalent representation:

$$\min_{\widehat{x},\widehat{c},\widehat{a},\widehat{b}} \sum_{i,j} f(C_{ij})(-\log(C_{ij}) - ||\widehat{x}_i - \widehat{c}_j||_2^2 + \widehat{a}_i + \widehat{b}_j))^2.$$

When combined with Lemma 1, we recognize this as a weighted multidimensional scaling objective with weights $f(C_{ij})$. Splitting the word vector $\widehat{x}_i$ and context vector $\widehat{c}_i$ is helpful in practice to optimize this objective, but not necessary under the assumptions of Lemma 1 since the true embedding $\widehat{x}_i = \widehat{c}_i = x_i/\sigma$ and $\widehat{a}_i, \widehat{b}_i = 0$ is a global minimum whenever $\dim(\widehat{x}) = d$. (See Thm S1.4 for detail)

---

[1] In practice, word embedding methods use a symmetrized window rather than counting transitions. This does not change any of the asymptotic analysis in the paper (Supplementary section S2)

**word2vec:** The embedding algorithm `word2vec` approximates a softmax objective:

$$\min_{\widehat{x},\widehat{c}} \sum_{i,j} C_{ij} \log\left(\frac{\exp(\langle \widehat{x}_i, \widehat{c}_j \rangle)}{\sum_{k=1}^{n} \exp(\langle \widehat{x}_i, \widehat{c}_k \rangle)}\right).$$

If $\dim(\widehat{x}) = d + 1$ we can set one of the dimensions of $\widehat{x} = 1$ as a bias term allowing us to rewrite the objective with a slack parameter $b_j$ analogously to `GloVe`. After reparametrization we obtain that for $\widehat{b} = b_j - ||\widehat{c}_j||_2^2$,

$$\min_{\widehat{x},\widehat{c},\widehat{b}} \sum_{i,j} C_{ij} \log\left(\frac{\exp(-||\widehat{x}_i - \widehat{c}_j||_2^2 + \widehat{b}_j)}{\sum_{k=1}^{n} \exp(-||\widehat{x}_i - \widehat{c}_k||_2^2 + \widehat{b}_k)}\right).$$

Since $C_{ij}/\sum_{k=1}^{n} C_{ik} \to \frac{\exp(-|||x_i - x_j||_2^2/\sigma^2)}{\sum_{k=1}^{n} \exp(-|||x_i - x_k||_2^2/\sigma^2)}$ this is the stochastic neighborhood embedding objective weighted by $\sum_{k=1}^{n} C_{ik}$. Once again, the true embedding $\widehat{x} = \widehat{c} = x/\sigma$ is a global minimum (Theorem S1.5). The negative sampling approximation used in practice behaves much like the SVD approach [9] and thus applying the same stationary point analysis as [9], the true embedding is a global minimum under the additional assumption that $||x_i||_2^2/\sigma = \log(\sum_j C_{ij}/\sqrt{\sum_{ij} C_{ij}})$.

**SVD:** The SVD approach [9] takes the log pointwise mutual information matrix:

$$M_{ij} = \log\left(C_{ij}\right) - \log\left(\sum_k C_{ik}\right) - \log\left(\sum_k C_{kj}\right) + \log\left(\sum_{ij} C_{ij}\right)$$

and applies the SVD to the shifted and truncated matrix : $(M_{ij} + \tau)_+$. This shift and truncation is done for computational reasons and to prevent $M_{ij}$ from diverging. In the limit where $m \to \infty$ and no truncation is performed there exists a shift $\tau$ as a function of $m$ such that the algorithm recovers the underlying embedding assuming $||x_i||_2^2/\sigma = \log(\sum_j C_{ij}/\sqrt{\sum_{ij} C_{ij}})$ (Lemma S1.3).

This assumption can be relaxed via a small modification to the algorithm: assume without loss of generality that the latent word vectors are mean-centered. Then we create the centered inner product matrix using the centering matrix $V = I - \mathbf{1}\mathbf{1}^T/n$

$$\widehat{M}_{ij} = V M_{ij} V^T / 2.$$

This is exactly classical multidimensional scaling and $\widehat{M}_{ij} \to \langle x_i, x_j \rangle/\sigma^2$ since the centering removes offsets $a_i, b_j$ and norms $||x_i||_2^2$ making SVD of $\widehat{M}_{ij}$ recover $x_i$ and $x_j$ (Theorem S1.2).

### 3.3 Metric regression from log co-occurences

We have demonstrated that by reparametrizing and taking on additional assumptions, existing word embedding algorithms could be cast as metric recovery under Lemma 1. However, it is not known if metric recovery would be effective in practice; for this we propose a new model which directly models Lemma 1 and acts as litmus test for our metric recovery paradigm.

Lemma 1 describes a log-linear relationship between distance and co-occurences. The canonical way to fit such a relationship would be to use a generalized linear model, where the co-occurences $C_{ij}$ follow a negative binomial distribution:

$$C_{ij} \sim \text{NegBin}\left(\theta, \theta(\theta + \exp(-||x_i - x_j||_2^2/2 + a_i + b_j))^{-1}\right).$$

Under this overdispersed log linear model, $\mathbb{E}[C_{ij}] = \exp(-||x_i - x_j||_2^2/2 + a_i + b_j)$, $\text{Var}(C_{ij}) = \mathbb{E}[C_{ij}]^2/\theta + \mathbb{E}[C_{ij}]$. Here, the parameter $\theta$ controls the contribution of large $C_{ij}$ and acts similarly to `GloVe`'s $f(C_{ij})$ weight function, which we cover in detail below. Fitting this model is straightforward, as we can define the log-likelihood in terms of the expected rate $\lambda_{ij} = \exp(-||x_i - x_j||_2^2/2 + a_i + b_j)$

$$\text{llh}(x, a, b, \theta) = \sum_{i,j} \theta \log(\theta) - \theta \log(\lambda_{ij} + \theta) + C_{ij} \log\left(1 - \frac{\theta}{\lambda_{ij} + \theta}\right) + \log\left(\frac{\Gamma(C_{ij} + \theta)}{\Gamma(\theta)\Gamma(C_{ij} + 1)}\right)$$

and perform gradient descent over the parameters, giving a simple update formula in terms of the error as

$$\delta_{ij} = \frac{(C_{ij} - \lambda_{ij})\theta}{\lambda_{ij} + \theta} \qquad dx_i = \sum_j (x_j - x_i)(\delta_{ij} + \delta_{ji}) \qquad da_i = \sum_j \delta_{ij} \qquad db_j = \sum_i \delta_{ij} \quad (2)$$

Optimizing this objective using stocahstic gradient descent will randomly select word pairs $i, j$ and attract or repulse the vectors $\widehat{x}$ and $\widehat{c}$ in order to achieve the relationship in Lemma 1. Our implementation uses the `GloVe` codebase (section S5.1 for details).

**Relationship to `GloVe`:** The overdispersion parameter $\theta$ sheds light on the role of `GloVe`'s weight function $f(C_{ij})$. Taking the Taylor expansion of the log-likelihood at $\log(\lambda_{ij}) \approx -\log(C_{ij})$ we find that for a constant $k_{ij}$,

$$\text{llh}(x, a, b, \theta) = \sum_{ij} k_{ij} - \frac{C_{ij}\theta}{2(C_{ij} + \theta)} (\log(\lambda_{ij}) - \log(C_{ij}))^2 + o((\log(\lambda_{ij}) - \log(C_{ij}))^3).$$

Note the similarity of the second order term with the `GloVe` objective. Both weight functions $\frac{C_{ij}\theta}{2(C_{ij} + \theta)}$ and $f(C_{ij}) = max(C_{ij}^{3/4}, x_{max}^{3/4})$ smoothly asymptote, downweighting large co-occurences. However, the empirical performance suggests that in practice, optimizing the distances directly and using the negative binomial loss consistently improves performance.

## 4 Metric recovery from Markov processes on graphs and manifolds

Metric recovery from random walks is possible under substantially more general conditions than the simple Markov process in Eq 1. We take an extreme view here and show that even a random walk over an *unweighted* directed graph holds enough information for metric recovery provided that the graph itself is suitably constructed in relation to the underlying metric.[2] To this end, we use a limiting argument (large vocabulary limit) with increasing numbers of points $\mathcal{X}_n = \{x_1, \ldots, x_n\}$, where $x_i$ are sampled i.i.d. from a density $p(x)$ over a compact Riemannian manifold. For our purposes, $p(x)$ should have a bounded log-gradient and a strict lower bound $p_0$ over the manifold. Since the points are assumed to lie on the manifold, we use the squared geodesic distance $\rho(x_i, x_j)^2$ in place of $\|x_i - x_j\|_2^2$ used earlier. The random walks we consider are over *unweighted spatial graphs* defined as

**Definition 2** (Spatial graph). *Let $\sigma_n : \mathcal{X}_n \to \mathbb{R}_{>0}$ be a local scale function and $h : \mathbb{R}_{\geq 0} \to [0, 1]$ a piecewise continuous function with sub-Gaussian tails. A spatial graph $G_n$ corresponding to $\sigma_n$ and $h$ is a random graph with vertex set $\mathcal{X}_n$ and a directed edge from $x_i$ to $x_j$ with probability $p_{ij} = h(\rho(x_i, x_j)^2/\sigma_n(x_i)^2)$.*

Simple examples of spatial graphs where the connectivity is not random ($p_{ij} = 0, 1$) include the $\varepsilon$ ball graph ($\sigma_n(x) = \varepsilon$) and the $k$-nearest neighbor graph ($\sigma_n(x) =$ distance to $k$-th neighbor) as in the $k$-nn graph, $\sigma_n$ is may depend on the set of points $\mathcal{X}_n$.

Our goal is to show that, as $n \to \infty$, we can recover $\rho(x_i, x_j)$ from co-occurrence counts generated from simple random walks over $G_n$. Log co-occurences and the geodesic will be connected in two steps. (1) we use known results to show that a simple random walk over the spatial graph, properly scaled, behaves similarly to a diffusion process; (2) the log-transition probability of a diffusion process will be related to the geodesic metric on a manifold.

**(1) The limiting random walk on a graph:** Just as the simple random walk over the integers converges to a Brownian motion, we may expect that under specific constraints the simple random walk $X_t^n$ over the graph $G_n$ will converge to some well-defined continuous process. We require that the scale functions converge to a continuous function $\bar{\sigma}$ ($\sigma_n(x) \xrightarrow{a.s.} g_n\bar{\sigma}(x)$); the size of a single step vanish ($g_n \to 0$) but contain at least a polynomial number of points within $\sigma_n(x)$ ($g_n n^{\frac{1}{d+2}} \log(n)^{-\frac{1}{d+2}} \to \infty$). Under this limit, our assumptions about the density $p(x)$, and an additional regularity condition, [3]

**Theorem 3** (Stroock-Varadhan on graphs[7, 22]). *The simple random walk $X_t^n$ on $G_n$ converges in Skorokhod space $\mathsf{D}([0, \infty), \overline{D})$ after a time scaling $\widehat{t} = tg_n^2$ to the Itô process $Y_{\widehat{t}}$ valued in $\mathsf{C}([0, \infty), \overline{D})$ as $X_{\widehat{t}g_n^{-2}}^n \to Y_{\widehat{t}}$. The process $Y_{\widehat{t}}$ is defined over the normal coordinates of the manifold $(D, g)$ with reflecting boundary conditions on $D$ as*

$$dY_{\widehat{t}} = \nabla \log(p(Y_{\widehat{t}}))\bar{\sigma}(Y_{\widehat{t}})^2 d\widehat{t} + \bar{\sigma}(Y_{\widehat{t}})dW_{\widehat{t}} \qquad (3)$$

---

[2]The weighted graph case follows identical arguments, replacing Theorem 3 with [22, Theorem 3].

[3]To ensure convergence of densities, require in addition that for $t = \Theta(g_n^{-2})$, the rescaled marginal distribution $n\mathbb{P}(X_t|X_0)$ is a.s. uniformly equicontinuous. For undirected spatial graphs, this is known to be true[4] for spatial graphs, but for directed graphs this is an open conjecture highlighted in [7]

**(2) Log transition probability as a metric** We may now use the stochastic process $Y_{\widehat{t}}$ to connect the log transition probability to the geodesic distance using Varadhan's large deviation formula.

**Theorem 4** (Varadhan [24, 15]). *Let $Y_t$ be a Itô process defined over a complete Riemann manifold $(D, g)$ with geodesic distance $\rho(x_i, x_j)$ then*

$$\lim_{t \to 0} -t \log(\mathbb{P}(Y_t = x_j | Y_0 = x_i)) \to \rho(x_i, x_j)^2.$$

This estimate holds more generally for any space admitting a diffusive stochastic process [20]. Taken together, we finally obtain Varadhan's formula over graphs:

**Corollary 5** (Varadhan's formula on graphs). *For any $\delta, \gamma, n_0$ there exists some $\widehat{t}$, $n > n_0$, and sequence $b_j^n$ such that the following holds for the simple random walk $X_t^n$:*

$$\mathbb{P}\Big(\sup_{x_i, x_j \in \mathcal{X}_{n_0}} \Big|\widehat{t} \log(\mathbb{P}(X_{\widehat{t}g_n^{-2}}^n = x_j \mid X_0^n = x_i)) - \widehat{t}b_j^n - \rho_{\overline{\sigma}(x)}(x_i, x_j)^2\Big| > \delta\Big) < \gamma$$

*Where $\rho_{\overline{\sigma}(x)}$ is the geodesic defined as $\rho_{\overline{\sigma}(x)}(x_i, x_j) = \min_{f \in C^1: f(0) = x_i, f(1) = x_j} \int_0^1 \overline{\sigma}(f(t)) dt$*

*Proof.* Sketch: For the Itô process, Varadhan's formula (Theorem 4) implies that we can find some time $\widehat{t}$ such that the log-marginal distribution of $\widehat{Y}$ is close to the geodesic. To convert this statement to our graph setting, we use the convergence of stochastic processes (Theorem 3) with equicontinuity of marginals to ensure that after $t = \widehat{t}g_n^{-2}$ steps, the transition probability over the graph converges to the marginal distribution of $\widehat{Y}$. Finally, compactness of the domain implies that log-marginals converge resulting in Varadhan's formula for graphs (see Corollary S3.2 for details). □

Since the co-occurence $C_{ij}$ has the limit $\log(C_{ij} / \sum_k C_{ik}) \to \mathbb{P}(X_{t+1}^n = x_j \mid X_0^n = x_i)$, this results in an analog of Lemma 1 in the manifold setting. Our proof demonstrates that regardless of the graph weights and manifold structure, in the large-sample small-time limit, log co-occurences faithfully capture the underlying metric structure of the data. While there has been ad-hoc attempts to apply word embeddings to graph random walks [18], this theorem demonstrates that embedding the log co-occurence is a principled method for graph metric recovery.

**Generalizing the Markov sentence model:** The spatial Markov random walk defined above has two flaws: first, cannot properly account for function words such as *the* since whenever the Markov chain transitions from a topic to a function word, it forgets the original topic. Second, since the unigram frequency of a word is the stationary distribution, frequent words are geometrically constrained to be close to all other words. Both of these assumptions can be relaxed by assuming that a latent spatial Markov chain, which we call the topic process $Y_t$, generates the observed sentence process $X_t$. This idea of a latent topic model underlying word embeddings has been explored [1]; our contributions are threefold: we contextualize this model as part of a metric embedding framework, provide a intuitive proof that directly applies Varadhan's formula, and relax some constraints on the distribution of $Y$ by taking the large vocabulary limit (see section S4.1).

|  | Google Analogies (cos) | | | Google Analogies ($L_2$) | | | SAT | |
| --- | --- | --- | --- | --- | --- | --- | --- | --- |
| Method | Sem. | Synt. | Total | Sem. | Synt. | Total | $L_2$ | Cosine |
| Regression | **78.4** | 70.8 | **73.7** | **75.5** | 70.9 | **72.6** | 39.2 | 37.8 |
| GloVE | 72.6 | 71.2 | 71.7 | 65.6 | 66.6 | 67.2 | 36.9 | 33.6 |
| SVD | 57.4 | 50.8 | 53.4 | 53.7 | 48.2 | 50.3 | 27.1 | 25.8 |
| Word2vec | 73.4 | **73.3** | 73.3 | 71.4 | **70.9** | 71.1 | **42.0** | **42.0** |

Table 1: Regression and Word2vec perform well on Google and SAT analogies.

The topic process $Y_t$ is defined over $\mathbb{R}^d$ by local jumps according to a smooth subgaussian kernel $h$ with $\int_x ||x||_2^2 h(x) dx = \sigma_0$, movement rate $\sigma^2$ and a log-differentible topic distribution $w(x)$ which defines the stationary distribution of the current topic over the latent semantic space. [4]

$$\mathbb{P}(Y_{t+1}|Y_t) = h(||Y_{t+1} - (Y_t + \nabla \log(w(Y_t))\sigma^2)||_2^2/\sigma^2) \qquad (4)$$

Given a topic $Y_t$, we assume the probability of observing a particular word decays exponentially with the semantic distance between the current topic and word scaled by $\overline{\sigma}$, as well as a non-metric frequency $\alpha$ which accounts for the frequency of function words such as *the* and *and*.

$$\mathbb{P}(X_t = x_i | Y_t = y) \propto \alpha_i \exp(-||x_i - y||^2/\overline{\sigma}^2).$$

Under this general model, we obtain a heat kernel estimate analogous to Cor 5, with constraints on the new scale parameter $\overline{\sigma}$ (Theorem S4.1),

$$\mathbb{P}(X_t = x_j | X_0 = x_i) \propto \frac{\alpha_i}{\pi(x_i)} w(x_i) \exp\left(-\frac{||x_j - x_i||_2^2}{2(\overline{\sigma}^2 + t\sigma^2\sigma_0^2)}\right) \left(1 + O(\sigma^2\sigma_0^2 t) + O(\overline{\sigma}^2)\right) + O(t^{-1/2}).$$

This allows word embedding algorithms to handle latent processes under the same small neighborhood ($\sigma \to 0$), large window $t \to \infty$ limit assuming that $\overline{\sigma}_0$ is small relative to the Hessian of $w(x)$ (See Theorem S4.1 for details).

## 5 Empirical validation

We experimentally validate two aspects of our word embedding theory: the semantic space hypothesis, and the manifold Varadhan's formula. Our goal is not to find the absolute best method and evaluation metric for word embeddings, which has been studied at detail [10]. Instead we will demonstrate that word embeddings based on metric recovery is competitive with existing state-of-the-art in both manifold learning and semantic induction tasks.

---

[4] We assume Euclidean, rather than arbitrary manifold since the additivity of vectors implied by analogical reasoning tasks require Euclidean embeddings

## 5.1 Semantic spaces in vector word representations

**Corpus and training:** We trained all methods on three different corpora: 2.4B tokens from Wikipedia, 6.4B tokens used to train `word2vec`, and 5.8B tokens combining Wikipedia with Gigaword5 emulating `GloVe`'s corpus (section S5.2 for details). We show performance for the `GloVe` corpus throughout but include all corpora in section S7. Word embeddings were generated on the top 100K words for each corpus using four methods: `word2vec`, `GloVe`, randomized SVD (referred to as SVD), and metric regression (referred to as regression). (see section S5.1). [5]

For fairness we fix the hyperparameter for metric regression at $\theta = 50$, developing and testing the code exclusively on the first 1GB subset of the wiki dataset. Vectors used in this paper represent the first run of our method on each full corpus. For open-vocabulary tasks, we restrict the set of answers to the top 30K words which improves performance while covering the majority of the questions.

**Solving analogies using survey data alone:** We demonstrate that embedding semantic similarity derived from survey data is sufficient for solving analogies by replicating a study by Rumelhart and Abrahamson. In this experiment, shown in Table 2, we take a free-association dataset [16] where words are vertices on a graph and edge weights $w_{ij}$ represent the number of times that word $j$ was considered most similar to word $i$ in a survey. We take this the largest connected component of 4845 words and 61570 weights and embed this weighted graph using stochastic neighborhood embedding (SNE) and Isomap for which squared edge distances are defined as $-\log(w_{ij}/\max_{kl}(w_{kl}))$. Solving the Google analogy questions [11] covered by the 4845 words using these vectors shows that Isomap combined with surveys can outperform the corpus based metric regression vectors on semantic, but not syntactic tasks; this is due to the fact that free-association surveys capture semantic, but not syntactic similarity between words. These results support both the semantic field hypothesis, and the exponential decay of semantic similarity with embedded distance.

**Analogies:** The results on the Google analogies shown in Table 1 demonstrate that our proposed framework of metric regression and naive vector addition ($L_2$) is competitive with the baseline of `word2vec` with cosine distance. The performance gap across methods is small and fluctuates across corpora, but metric regression consistently outperforms `GloVe` on most tasks and outperforms all methods on semantic analogies, while `word2vec` does better on syntactic categories.

We also evaluate the methods on more difficult SAT type questions [23] where a prototype pair A:B is given and we must choose amongst a set of candidate pairs $[C_1 : D_1] \ldots [C_5 : D_5]$. In this evaluation, cosine similarity between vector differences is no longer the optimal choice and $L_2$ metric performs slightly better. In terms of methods, we find that `word2vec` is best, followed by metric regression. The results on these two analogy datasets show that directly embedding the log-coocurrence metric and taking $L_2$ distances between vectors is competitive with current approaches to analogical reasoning. The consistent improvement of metric embedding over `GloVe` despite their similarities in implementation (Section S5.1), parameters, and stationary point (Section 3.3) suggest that the metric embedding approach to word embedding can lead to algorithmic improvements.

---
[5] We used randomized, rather than full SVD due to the difficulty of scaling SVD to this problem size. For perfomance of full SVD factorizations see [10].

**Sequence and classification tasks:** We propose two new difficult inductive reasoning tasks based upon the semantic field hypothesis [21]. The sequence and classification datasets, as described in Section 2 are tasks that require one to pick either a sequence completion ($hour, minute, \ldots$) or find an element within the same category out of five possible choices. The questions were generated using WordNet semantic relations [13]. These datasets were constructed before any embeddings to avoid biasing them towards any one method (Section S5.3 for further details). As predicted by the semantic field hypothesis, word embeddings solve both tasks effectively, with metric embedding consistently performing well on these multiple choice tasks (Table 3).

The metric recovery approach of metric regression methods and $L_2$ distance can consistently perform as well as the current state-of-the-art on the three semantic tasks: Google semantic analogies, sequence, and classification.

## 5.2 Word embeddings can embed manifolds

**MNIST digits:** We evaluate whether word embeddings can perform nonlinear dimensionality reduction by embedding the MNIST digits dataset. Using a four-thousand point subset, we generated a k-nearest neighbor graph ($k = 20$) and generated 10 simple random walks of length 200 from each point resulting in 40,000 sentences each of length 200. We compared the four word embedding methods against standard dimensionality reduction methods: PCA, Isomap, SNE and, $t$-SNE. The quality of an embedding was measured using the percentage of 5-nearest neighbors having the same cluster label. The four embeddings shown in Fig. 2 demonstrate that metric regression is highly effective at this task, outperforming metric SNE and beaten only by $t$-SNE (91%), which is a visualization method designed for cluster separation. All word embedding methods including SVD (68%) embed the MNIST digits well and outperform baselines of PCA (48%) and Isomap (49%) (Suppplementary Figure S1). This empirically verifies the theoretical predictions in Corollary 5 that log co-occurences of a simple random walk converge to the squared geodesic.

## 6 Discussion

Our work further justifies word embeddings by linking them to semantic spaces from psychometric literature. The key conceptual glue is metric recovery from co-occurrences. The notion of semantic space, as well as our theoretical recovery results, suggest the $L_2$ distance can serve as a natural semantic metric. This is reasonably supported by our empirical analysis, including the consistent perfor-mance of the proposed direct regression method and the utility of $L_2$ distance in selecting analogies.

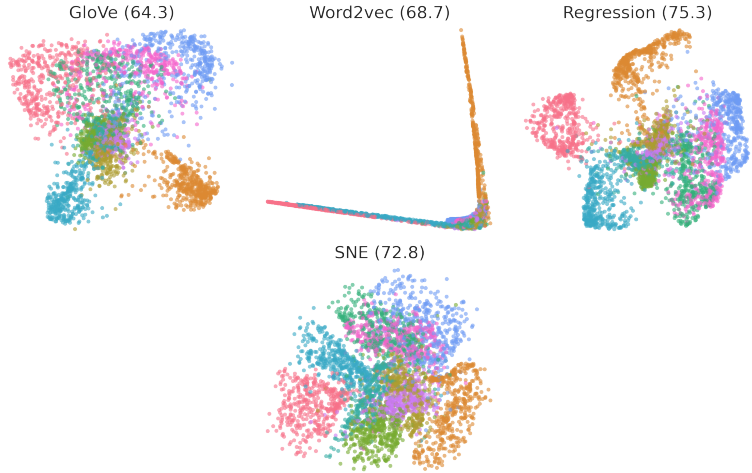

Figure 2: MNIST digit embedding using word embedding methods (left three) and metric embedding on the same graph (right). Performance is quantified by percentage of 5-nearest neighbors sharing the same cluster label.

|  | Manifold Learning | | Word Embedding |
|---|---|---|---|
| Analogy | Isomap | SNE | Regression |
| Semantic | **83.3** | 21.5 | 70.7 |
| Syntactic | 8.2 | 1.5 | **76.9** |
| Total | 51.4 | 13.1 | **73.4** |

Table 2: Word embedding generated using human semantic similarity surveys and manifold learning outperforms word embeddings from a corpus.

| | Classification | | Sequence | |
|---|---|---|---|---|
| Method | Cosine | $L_2$ | Cosine | $L_2$ |
| Regression | **84.6** | **87.6** | **59.0** | **58.3** |
| GloVE | 80.1 | 73.1 | **59.0** | 48.8 |
| SVD | 74.6 | 65.2 | 53.0 | 52.4 |
| Word2vec | **84.6** | 76.4 | 56.2 | 54.4 |

Table 3: Regression with $L_2$ loss performs well on semantic classification and sequence data

Our framework highlights the strong interplay between methods for learning word embeddings and manifold learning, suggesting several avenues for recovering vector representations of phrases and sentences via properly defined Markov processes and their generalizations.

# Supplement for: Word, graph and manifold embedding from Markov processes

September 18, 2015

## 1 Consistency of the global minima of word embedding algorithms

**Lemma S1.1** (Law of large numbers for log coocurrences). *Let $X_t$ be a Markov chain defined by the transition*

$$\mathbb{P}(X_t = x_j | X_{t-1} = x_i) = \frac{\exp(-||x_i - x_j||_2^2/\sigma^2)}{\sum_{k=1}^n \exp(-||x_i - x_k||_2^2/\sigma^2)} \quad (1)$$

*and $C_{ij}$ be the number of times that $X_t = x_j$ and $X_{t-1} = x_i$ over $m$ steps of this chain. Then for any $\delta > 0$ and $\varepsilon > 0$ there exist some $m$ and constants $a_i^m$ and $b_j^m$ such that*

$$\mathbb{P}\left(\sup_{i,j} \left| -\log(C_{ij}) - ||x_i - x_j||_2^2/\sigma^2 + a_i^m + b_j^m \right| > \delta \right) < \varepsilon$$

*Proof.* By detailed balance we observe that the stationary distribution $\pi_X(x_i)$ exists and is the normalization constant of the transition

$$\mathbb{P}(X_t = x_j | X_{t-1} = x_i)\pi_X(x_i) = \frac{\exp(-||x_i - x_j||_2^2/\sigma^2)}{\sum_{k=1}^n \exp(-||x_i - x_k||_2^2/\sigma^2)} \sum_{k=1}^n \exp(-||x_i - x_k||_2^2/\sigma^2)$$

$$= \mathbb{P}(X_t = x_i | X_{t-1} = x_j)\pi_X(x_j).$$

Define $m_i$ as the number of times that $X_t = x_i$ in a $m$ word corpus. Applying the Markov chain law of large numbers, we obtain that for any $\varepsilon_0 > 0$ and $\delta_0 > 0$ there exists some $m$ such that

$$P\left(\sup_i \left|\pi_X(x_i) - m_i/m\right| > \delta_0\right) < \varepsilon_0.$$

Therefore with probability $\varepsilon_0$, $m_i > m(\pi_X(x_i) - \delta_0)$.

Now given $m_i$, $C_{ij} \sim \text{Binom}(\mathbb{P}(X_t = x_j | X_{t-1} = X_i), m_i)$ applying Hoeffding's inequality and union bounding for any $\delta_1 > 0$ and $\varepsilon_1 > 0$ there exists some set of $m_i$ such that

$$\mathbb{P}\left(\sup_{i,j}\left|C_{ij}/m_i - \mathbb{P}\left(X_t = x_j \mid X_{t-1} = x_i\right)\right| < \delta_1\right) \geq (1 - 2\exp(-2\delta_1^2 m_i))^{n^2} = \varepsilon_1.$$

Since $||x_i - x_j||_2 < \infty$, $\mathbb{P}(X_t = x_j | X_{t-1} = x_i)$ is lower bounded by some strictly positive constant $c$ and we may apply the continuous mapping theorem on $\log(c)$ uniformly continuous over $(c, \infty)$ to obtain that for all $\delta_2$ and $\varepsilon_2$ there exists some set of $m_i$ such that

$$\mathbb{P}\left(\sup_{i,j}\left|\log(C_{ij}) - \log(m_i) - \log(\mathbb{P}(X_t = x_j | X_{t-1} = x_i))\right| < \delta_2\right) \geq \varepsilon_2.$$

Therefore given any $\delta$ and $\varepsilon$ for the theorem statement, set $\delta_2 = \delta$ and $\varepsilon_2 = \sqrt{\varepsilon}$ and define $m'$ as the smallest $m_i$ required. Since $\sup_{ij} ||x_i - x_j|| < \infty$, the Markov chain law of large numbers implies we can always find some $m$ such that $\inf_i m_i > m'$ with probability at least $\sqrt{\varepsilon}$ which completes the original statement. □

**Theorem S1.2** (Consistency of SVD-MDS). *Let $C_{ij}$ be defined as above and $M_{ij} = \log(C_{ij})$ and the centering matrix $V = I - 11^T/n$. Define the SVD based embedding $\widehat{X}$ as*

$$\widehat{X}\widehat{X}^T = \widehat{M} = VMV/2.$$

*Without loss of generality, also assume that the latent vectors $x$ have zero mean, then for any $\varepsilon > 0$ and $\delta > 0$, there exists some $m$, scaling constant $\sigma$, and an orthogonal matrix $A$ such that*

$$\mathbb{P}(\sum_i ||A\widehat{x}_i/\sigma^2 - x_j||_2^2 > \delta) < \varepsilon$$

*Proof.* By Lemma S1.1 we have that

$$P\left(\sup_{i,j}\left|-\log(C_{ij}) - ||x_i - x_j||_2^2/\sigma^2 + a_i^m + b_j^m\right| > \widehat{\delta}\right) < \widehat{\varepsilon}$$

Since mean error cannot exceed entrywise error we can bound the row averages of $\log(C_{ij})$, where the dot product term is zero since $x$ is zero mean.

$$P\left(\sup_i\left|-\frac{\sum_j \log(C_{ij})}{n} - a_i^m - \frac{\sum_j b_j^m}{n} - \frac{||x_i||_2^2}{\sigma} - \frac{\sum_j ||x_j||_2^2}{\sigma^2 n} + 2\left\langle x_i, \frac{\sum_j x_j}{n}\right\rangle\right| > \widehat{\delta}\right) < \widehat{\varepsilon}$$

Or in other words, $-\frac{\sum_j \log(C_{ij})}{n} \approx a_i^m + \frac{\sum_j b_j^m}{n} - \frac{||x_i||_2^2}{\sigma} - \frac{\sum_j ||x_j||_2^2}{\sigma^2 n}$

Define $M'_{ij} = -\log(C)_{ij} - \frac{\sum_j -\log(C)_{ij}}{n}$; applying the triangle inequality and combining both bounds gives

$$P\left(\sup_j\left|\frac{\sum_i M'_{ij}}{n} - \left(b_j - \frac{\sum_k b_k^m}{n} + ||x_j|| - \frac{\sum_k ||x_k||_2^2}{\sigma^2 n}\right)\right| > 2\widehat{\delta}\right) < 1 - (1-\widehat{\varepsilon})^2.$$

Note that $M'_{ij} - \sum_i M'_{ij}/n = 2\widehat{M}_{ij}$ is the doubly centered matrix as defined above and combining all above bounds we have,

$$P\left(\sup_{ij}\left|\widehat{M}_{ij} - \langle x_i, x_j \rangle\right| > 4\widehat{\delta}\right) < 1 - (1-\widehat{\varepsilon})^4.$$

Given that the dot product matrix has error at most $4\delta$ the resulting embedding it known to have at most $\sqrt{4\widehat{\delta}}$ error [15].

This completes the proof, since we can pick $\widehat{\delta} = \delta^2/4$ and $\widehat{\varepsilon} = 1 - (1-\varepsilon)^{1/4}$ □

**Lemma S1.3** (Consistency of SVD). *Assume the conditions of Theorem S1.2 and additionally, assume the norm of the latent embedding is proportional to the unigram frequency*

$$||x_i||/\sigma^2 = \frac{\sum_j C_{ij}}{\sqrt{\sum_{ij} C_{ij}}}.$$

*Under these conditions, Let $\widehat{X}$ be the embedding derived from the SVD of $M_{ij}$ as*

$$2\widehat{X}\widehat{X}^T = M_{ij} = \log(C_{ij}) - \log\left(\sum_k C_{ik}\right) - \log\left(\sum_k C_{kj}\right) + \log\left(\sum_{ij} C_{ij}\right) + \tau.$$

*Then there exists a $\tau$ such that this embedding is close to the true embedding under the same equivalence class as Lemma S1.3*

$$\mathbb{P}\Big(\sum_i ||A\widehat{x}_i/\sigma^2 - x_j||_2^2 > \delta\Big) < \varepsilon.$$

*Proof.* By Lemma S1.1, for any $\delta_0 > 0$ and $\varepsilon_0 > 0$ there exists a $m$ such that

$$P\left(\sup_{i,j}\left|-\log(C_{ij}) - \|x_i - x_j\|_2^2/\sigma^2 + \log\left(\sum_{k=1}^n \exp(-\|x_i - x_k\|_2^2/\sigma^2)\right) - \log\left(\sum_k C_{ik}\right)\right| > \delta_0\right) < \varepsilon_0$$

which implies that for any $\delta_1 > 0$ and $\varepsilon_1 > 0$ there exists a $m$ such that

$$P\left(\sup_{i,j}\left|-\log(C_{ij}) - (\|x_i - x_j\|_2^2/\sigma^2) - \log(mc)\right| > \delta_1\right) < \varepsilon_1.$$

Now additionally, if $\sum_k C_{ik}/\sqrt{\sum_{ij} C_{ij}} = \|x_i\|^2/\sigma^2$ then we can rewrite the above bound as

$$P\left(\sup_{i,j}\left|\log(C_{ij}) - \log\left(\sum_k C_{ik}\right) - \log\left(\sum_k C_{kj}\right) + \log\left(\sum_{ij} C_{ij}\right) - 2\langle x_i, x_j\rangle/\sigma^2 - \log(mc)\right| > \delta_1\right) < \varepsilon_1.$$

and therefore,

$$P\left(\sup_{i,j}\left|M_{ij} - 2\langle x_i, x_j\rangle/\sigma^2 - \log(mc)\right| > \delta_1\right) < \varepsilon_1.$$

Given that the dot product matrix has error at most $\delta_1$, the resulting embedding it known to have at most $\sqrt{\delta_1}$ error [15].

This completes the proof, since we can pick $\tau = -\log(mc)$, $\delta_1 = \delta^2$ and $\varepsilon_1 = \varepsilon$. □

**Theorem S1.4** (Consistency of GloVE). *Define the GloVe objective function as*

$$g(\widehat{x}, \widehat{c}, \widehat{a}, \widehat{b}) = \sum_{i,j} f(C_{ij})(2\widehat{x}_i\widehat{c}_j + \widehat{a} + \widehat{b} - \log(C_{ij}))^2$$

*Define $\bar{x}_m, \bar{c}_m, \bar{a}_m, \bar{b}_m$ as the global minima of the above objective function for a corpus of size $m$.*

*Then the parameters derived from the true embedding in Lemma S1.1, $x' = x/\sigma$, $a'_i = a_i^m - \|x_i\|_2^2/\sigma^2$, $b'_i = b_i^m - \|x_i\|_2^2/\sigma^2$ is arbitrarily close to the global minima in the sense that for any $\varepsilon > 0$ and $\delta > 0$ there exists some $m$ such that*

$$\mathbb{P}(|g(x', x', a', b') - g(\bar{x}_m, \bar{c}_m, \bar{a}_m, \bar{b}_m)| > \delta) < \varepsilon$$

*Proof.* Using Lemma S1.1 with error $\delta_0$ and probability $\varepsilon_0$ there exists some $m$ such that uniformly over $i$ and $j$,

$$(-\|x_i - x_j\|_2^2/\sigma^2 + a_i^m + b_i^m + \log(C_{ij}))^2 \leq \delta_0^2.$$

Now recall that $f(C_{ij}) \leq 10^{3/4} = c$ therefore

$$\mathbb{P}(g(x', x', a', b') > cn^2\delta_0^2) < \varepsilon_0.$$

Now the global minima $g(\bar{x}_m, \bar{c}_m, \bar{a}_m, \bar{b}_m)$ must be less than $g(x', x', a', b')$ and we have $0 < g(\bar{x}_m, \bar{c}_m, \bar{a}_m, \bar{b}_m) < g(x', x', a', b')$.

Therefore,

$$\mathbb{P}(|g(x', x', a', b') - g(\bar{x}_m, \bar{c}_m, \bar{a}_m, \bar{b}_m)| > cn^2\delta_0/2) < \varepsilon_0.$$

Picking a $m$ such that $\delta_0 = 2\delta/(cn^2)$ and $\varepsilon_0 = \varepsilon$ concludes the proof. □

**Theorem S1.5** (Consistency of softmax/word2vec). *Define the softmax objective function with bias as*

$$g(\widehat{x}, \widehat{c}, \widehat{b}) = \sum_{ij} C_{ij} \log\left(\frac{\exp(-\|\widehat{x}_i - \widehat{c}_j\|_2^2 + \widehat{b}_j)}{\sum_{k=1}^n \exp(-\|\widehat{x}_i - \widehat{c}_k\|_2^2 + \widehat{b}_k)}\right)$$

*Define $\bar{x}_m, \bar{c}_m, \bar{b}_m$ as the global minima of the above objective function for a corpus of size $m$. We claim that for any $\varepsilon > 0$ and $\delta > 0$ there exists some $m$ such that*

$$\mathbb{P}(|g(x/\sigma, x/\sigma, 0) - g(\bar{x}, \bar{c}, \bar{b})| > \delta) < \varepsilon$$

*Proof.* By differentiation, any objective of the form

$$\min_{\lambda_{ij}} C_{ij} \log\left(\frac{\exp(-\lambda_{ij})}{\sum_k \exp(-\lambda_{ik})}\right)$$

has the minima $\lambda_{ij} = -\log(C_{ij}) + a_i$ with objective function value $C_{ij}\log(C_{ij}/\sum_k C_{ik})$. This gives a global function lower bound

$$g(\overline{x}, \overline{c}, \overline{b}) \geq \sum_{ij} C_{ij} \log\left(\frac{C_{ij}}{\sum_k C_{ik}}\right)$$

Now consider the function value of the true embedding $x/\sigma$;

$$g(x/\sigma, x/\sigma, 0) = \sum_{ij} C_{ij} \log\left(\frac{\exp(-\|x_i - x_j\|_2^2/\sigma^2)}{\sum_k \exp(-\|x_i - x_k\|_2^2/\sigma^2)}\right)$$

$$= \sum_{ij} C_{ij} \log\left(\frac{\exp(\log(C_{ij}) + \delta_{ij} + a_i)}{\sum_k \exp(\log(C_{ik}) + \delta_{ik} + a_i)}\right).$$

We can bound the error variables $\delta_{ij}$ using Lemma S1.1 as $\sup_{ij} |\delta_{ij}| < \delta_0$ with probability $\varepsilon_0$ for sufficiently large $m$ with $a_i = \log(m_i) - \log(\sum_{k=1}^n \exp(-\|x_i - x_k\|_2^2/\sigma^2))$.

Taking the Taylor expansion at $\delta_{ij} = 0$, we have

$$g(x/\sigma, x/\sigma, 0) = \sum_{ij} C_{ij} \log\left(\frac{C_{ij}}{\sum_k C_{ik}}\right) + \sum_{l=1}^n \frac{C_{il}}{\sum_k C_{ik}} \delta_{il} + o(\|\delta\|_2^2)$$

Applying Lemma S1.1 we obtain:

$$\mathbb{P}\left(\left|g(x/\sigma, x/\sigma, 0) - \sum_{ij} C_{ij} \log\left(\frac{C_{ij}}{\sum_k C_{ik}}\right)\right| > n\delta_0\right) < \varepsilon_0$$

Combining with the global function lower bound we have that

$$\mathbb{P}\left(\left|g(x/\sigma, x/\sigma, 0) - g(\overline{x}, \overline{c}, \overline{b})\right| > n\delta_0\right) < \varepsilon_0.$$

To obtain the original theorem statement, take $m$ to fulfil $\delta_0 = \delta/n$ and $\varepsilon_0 = \varepsilon$. □

Note that for negative-sampling based word2vec, applying the stationary point analysis of [7] combined with the analysis in Lemma S1.3 shows that the true embedding is a global minima.

**Theorem S1.6** (Metric regression consistency). *Define the negative binomial objective function*

$$\lambda(\widehat{x}, \widehat{c}, \widehat{a}, \widehat{b}) = \exp(-\|\widehat{x}_i - \widehat{c}_j\|_2^2/2 + a_i + b_j)$$

$$g(\widehat{x}, \widehat{c}, \widehat{a}, \widehat{b}, \theta) = \sum_{i,j} \theta \log(\theta) - \theta \log(\lambda(\widehat{x}_i, \widehat{c}_j, \widehat{a}_i, \widehat{b}_j) + \theta) + C_{ij} \log\left(1 - \frac{\theta}{\lambda(\widehat{x}_i, \widehat{c}_j, \widehat{a}_i, \widehat{b}_j) + \theta}\right) + \log\left(\frac{\Gamma(C_{ij} + \theta)}{\Gamma(\theta)\Gamma(C_{ij} + 1)}\right)$$

*Then the parameters derived from the true embedding in Lemma S1.1, $x' = x/\sigma$, $a'_i = a^m$, $b'_i = b_i^m$ is arbitrarily close to the global minima $g(\overline{x}_m, \overline{c}_m, \overline{a}_m, \overline{b}_m)$ in the sense that for any $\varepsilon > 0$ and $\delta > 0$ there exists some $m$ such that*

$$\mathbb{P}(|g(x', x', a', b') - g(\overline{x}_m, \overline{c}_m, \overline{a}_m, \overline{b}_m)| > \delta) < \varepsilon$$

*Proof.* The proof proceeds identically to that of Theorem S1.5. First obtain the global minima at $\lambda(\widehat{x}_i, \widehat{c}_j, \widehat{a}_i, \widehat{b}_j) = C_{ij}$) as

$$g(\overline{x}_m, \overline{c}_m, \overline{a}_m, \overline{b}_m) \geq \sum_{ij} k_{ij}$$

where

$$k_{ij} = C_{ij} \left(\log(C_{ij}) - \log(C_{ij} + \theta) + \theta(\log(\theta) - \log(C_{ij} + \theta)) + \log(\Gamma(C_{ij} + \theta)) - \log(\Gamma(\theta)) - \log(C_{ij} + 1)\right).$$

As with Theorem S1.5, rewriting $\lambda(\widehat{x}_i, \widehat{c}_j, \widehat{a}_i, \widehat{b}_j) = C_{ij} \exp(\delta_{ij})$ allows us to take the taylor expansion for $\exp(\delta_{ij})$ small, giving

$$llh(x, a, b, \theta) = \sum_{ij} k_{ij} - \frac{C_{ij}\theta}{2(C_{ij} + \theta)} (\delta_{ij})^2 + o(\delta_{ij}^3).$$

Applying Lemma S1.1 we obtain:

$$\mathbb{P}\left(\left|g(x', x', a', b') - \sum_{ij} k_{ij}\right| > n\delta_0\right) < \varepsilon_0$$

which when combined with the global function bound yields that the global minima is consistent. □

## 2 Symmetry and windowing co-occurences

Existing word embedding algorithms utilize weighted, windowed, symmetrized word counts. Let $C_{ij}^t$ define the $t$-step co-occurence which counts the number of times $X_{t+t'} = x_j$ and $X_{t'} = x_i$.

Then for some weight function $w(t)$ such that $\sum_{t=1}^{\infty} w(t) = 1$, we define

$$\widehat{C}_{ij} = \sum_{t=1}^{\infty} w(t)(C_{ij}^t + C_{ji}^t).$$

This is distinct from our stochastic process approach in two ways: first, there is symmetrization by counting both forward and backward transitions of the Markov chain. second, all words within a window of the center word $X_{t'}$ are used to form the co-occurences.

**Symmetry:** We begin by considering asymmetry of the random walk. If the Markov chain is reversible as in the cases of the Gaussian random walk, un-directed graphs, and the topic model, we can apply detailed balance to show that the joint distributions are symmetric

$$\mathbb{P}(X_{t+1} = x_j | X_t = x_i)\pi_X(x_i) = \mathbb{P}(X_{t+1} = x_i | X_t = x_j)\pi_X(x_j)$$

Therefore the empirical sum converges to

$$C_{ij}^t + C_{ji}^t \to \mathbb{P}(X_{t+t'} = x_j, X_{t'} = x_i) + \mathbb{P}(X_{t+t'} = x_i, X_{t'} = x_j) = 2\mathbb{P}(X_{t+t'} = x_j, X_{t'} = x_i)$$

In the cases where the random walk is non-reversible, such as a $k$-nearest neighbor graph then the two terms are not exactly equal, however note that if the non-symmetrized transition matricies $C_{ij}$ fulfill Varadhan's formula both ways:

$$-t\log(C_{ij}) - a_i^m \to ||x_i - x_j||_2^2 + b_j^m \qquad \text{and} \qquad -t\log(C_{ji}) - a_j^m \to ||x_j - x_i||_2^2 + b_i^m$$

The sum $\widehat{C}_{ij}$ will fulfil

$$(C_{ij}^t + C_{ji}^t) = \exp(-||x_i - x_j||_2^2/t + o(1/t)) \left(\exp(a_i/t + b_j/t) + \exp(b_i/t + a_j/t)\right)$$

and

$$-t\log(C_{ij}^t + C_{ji}^t) = ||x_i - x_j||_2^2 + \log\left(\exp(a_i/t + b_j/t) + \exp(b_i/t + a_j/t)\right) t + o(1)$$

More specifically, for the manifold case, $a_i = \log(\pi_{X_n}) \to \log(np(x)/\overline{\sigma}(x_i)^2)$ and $b_j = -\log(np(x))$, and so the above term reduces to

$$-t\log(C_{ij}^t + C_{ji}^t) = ||x_i - x_j||_2^2 + \log\left(\overline{\sigma}^{-2}(x_i) + \overline{\sigma}^{-2}(x_j)\right) t + o(1)$$

Since the $\overline{\sigma}$ is independent of $t$, as $t \to 0$, we are once again left with Varadhan's formula in the symmetrized case.

In practice, this does not seem to affect the manifold embedding approaches much; in the results section we attempt embedding the MNIST digits dataset using the $k$-nearest neighbor simple random walk which is nonreversible.

**Windowing:** Now we consider the effect of windowing. We focus on the manifold case for analytic simplicity, but the same limits apply to the other two examples of Gaussian random walks and topic models.

Let $q_t(x, x') = \mathbb{P}(Y_t = x | Y_0 = x')$ and where $Y_t$ fulfills Varadhan's formula such that there exists a metric function $\rho$,

$$\lim_{t \to 0} -t \log(q_t(x, x')) \to \rho(x, x')^2$$

Under these conditions, let $\widehat{q}_t(x, x') = \int_0^t q_{t'}(x, x')/t \, dt'$ define the windowed marginal distribution. We show this follows a windowed Varadhan's formula.

$$\lim_{t \to 0} t \widehat{q}_t(x, x') \to \rho(x, x')^2$$

This can be done via a direct argument. Varadhan's formula implies that,

$$q_t(x, x') = \exp\left(-\frac{\rho(x, x')^2}{t} + o\left(\frac{1}{t}\right)\right).$$

Thus we can find some bounding constants $0 < c = o(1)$ such that

$$\int_0^t \frac{1}{t} \exp\left(-\frac{\rho(x, x')^2}{t'} - \frac{c}{t'}\right) dt \le \widehat{q}_t(x, x') \le \int_0^t \frac{1}{t} \exp\left(-\frac{\rho(x, x')^2}{t'} + \frac{c}{t'}\right) dt'$$

Performing the bounding integral for general $c \in \mathbb{R}$,

$$\int_0^t \frac{1}{t} \exp\left(-\frac{\rho(x, x')^2}{t'} + \frac{c}{t'}\right) dt' = \frac{1}{t}\left(\exp\left(-\frac{\rho(x, x')^2 - 2c}{2t}\right) t + (c - \rho(x, x')^2/2)\Gamma\left(\frac{\rho(x, x')^2 - 2c}{2t}\right)\right)$$

$$= \frac{1}{t}\left(\exp\left(-\frac{c}{t} - \frac{\rho(x, x')^2}{t}\right)\left(-\frac{2t}{2c - \rho(x, x')^2} + t^2\right)\right)$$

Therefore we have that for any $c$,

$$\lim_{t \to 0} -t \log\left(\int_0^t \frac{1}{t} \exp\left(-\frac{\rho(x, x')^2}{t'} + \frac{c}{t'}\right) dt'\right) \to \rho(x, x')^2 - c$$

By the two-sided bound and $c = o(1)$,

$$\lim_{t \to 0} t \widehat{q}_t(x, x') \to \rho(x', x)^2.$$

as desired.

## 3  Varadhan's formula on graphs

We first prove the convergence of marginal densities under the assumption of equicontinuity.

**Lemma S3.1** (Convergence of marginal densities). *Let $x_0$ be some point in our domain $\mathcal{X}_n$ and define the marginal densities*

$$\widehat{q}_t(x) = \mathbb{P}(Y_t = x | Y_0 = x_0)$$
$$q_{t_n}(x) = \mathbb{P}(X_t^n = x | X_0^n = x_0)$$

*If $t_n g_n^2 = \widehat{t} = \Theta(1)$, then under condition $(\star)$ and the results of Theorem 3 such that $X_t^n \to Y_t^n$ weakly, we have*

$$\lim_{n \to \infty} n q_{t_n}(x) = \frac{\widehat{q}_{\widehat{t}}(x)}{p(x)}.$$

*Proof.* The a.s. weak convergence of processes of Theorem 3 implies by [2, Theorem 4.9.12] that the empirical marginal distribution

$$d\mu_n = \sum_{i=1}^n q_{t_n}(x_i)\delta_{x_i}$$

converges weakly to its continuous equivalent $d\mu = \widehat{q}_{\widehat{t}}(x)dx$ for $Y_{\widehat{t}}$. For any $x \in \mathcal{X}$ and $\delta > 0$, weak convergence against the test function $1_{B(x,\delta)}$ yields

$$\sum_{y \in \mathcal{X}_n, |y-x|<\delta} q_{t_n}(y) \to \int_{|y-x|<\delta} \widehat{q}_{\widehat{t}}(y)dy.$$

By uniform equicontinuity of $nq_t(x)$, for any $\varepsilon > 0$ there is small enough $\delta > 0$ so that for all $n$ we have

$$\left| \sum_{y \in \mathcal{X}_n, |y-x|<\delta} q_{t_n}(y) - |\mathcal{X}_n \cap B(x,\delta)|\widehat{q}_t(x) \right| \leq n^{-1}|\mathcal{X}_n \cap B(x,\delta)|\varepsilon,$$

which implies that

$$\lim_{n \to \infty} q_{t_n}(x)p(x)n = \lim_{\delta \to 0} \lim_{n \to \infty} V_d^{-1}\delta^{-d} nq_{t_n}(x) \int_{|y-x|<\delta} p(y)dy$$

$$= \lim_{\delta \to 0} \lim_{n \to \infty} V_d^{-1}\delta^{-d}|\mathcal{X}_n \cap B(x,\delta)|q_{t_n}(x) = \lim_{\delta \to 0} V_d^{-1}\delta^{-d} \int_{|y-x|<\delta} \widehat{q}_{\widehat{t}}(y)dy = \widehat{q}_{\widehat{t}}(x).$$

We conclude the desired

$$\lim_{n \to \infty} nq_t(x) = \frac{\widehat{q}_{\widehat{t}}(x)}{p(x)}.\qquad\square$$

Given this, we can now prove Varadhan's formula specialized to the manifold graph case:

**Corollary S3.2** (Heat kernel estimates on graphs). *For any $\delta > 0, \gamma > 0, n_0 > 0$ there exists some $\widehat{t}$, $n > n_0$, and sequence $b_j^n$ such that the following holds for the simple random walk $X_t^n$:*

$$\mathbb{P}\left( \sup_{x_i, x_j \in \mathcal{X}_{n_0}} \left| \widehat{t} \log(\mathbb{P}(X_{\widehat{t}g_n^{-2}}^n = x_j \mid X_0^n = x_i)) - \widehat{t}b_j^n - \rho_{\overline{\sigma}(x)}(x_i, x_j)^2 \right| > \delta \right) < \gamma$$

*Where $\rho_{\overline{\sigma}(x)}$ is the geodesic defined by $\overline{\sigma}(x)$:*

$$\rho_{\overline{\sigma}(x)}(x_i, x_j) = \min_{f \in C^1: f(0)=x_i, f(1)=x_j} \int_0^1 \overline{\sigma}(f(t))dt$$

*Proof.* The proof is in two parts. First, by Varadhan's formula (Theorem 4, [13, Eq. 1.7]) for any $\delta_1 > 0$ there exists some $\widehat{t}$ such that:

$$\sup_{y,y' \in D} |-\widehat{t}\log(\mathbb{P}(Y_{\widehat{t}} = y'|Y_0 = y)) - \rho_{\overline{\sigma}(x)}(y',y)^2| < \delta_1$$

Now uniform equicontinuity of marginals implies uniform convergence of marginals (Lemma S3.1) and therefore for any $\delta_2 > 0$ and $\gamma_0$, there exists a $n$ such that,

$$\mathbb{P}(\sup_{x_j, x_i \in \mathcal{X}_{n_0}} |\mathbb{P}(Y_{\widehat{t}} = x_j|Y_0 = x_i) - np(x_j)\mathbb{P}(X_{g_n^{-2}\widehat{t}}^n = x_j|X_0^n = x_i)| > \delta_2) < \gamma_0$$

By the lower bound on $p$ and compactness of the domain $D$, $\mathbb{P}(Y_{\widehat{t}}|Y_0)$ is lower bounded by some strictly positive constant $c$ and we can apply uniform continuity of $\log(x)$ over $(c, \infty)$ to get that for some $\delta_3$ and $\gamma$,

$$\mathbb{P}(\sup_{x_j, x_i \in \mathcal{X}_{n_0}} |\log(\mathbb{P}(Y_{\widehat{t}} = x_j|Y_0 = x_i)) - \log(np(x_j)) - \log(\mathbb{P}(X_{g_n^{-2}\widehat{t}}^n = x_j|X_0^n = x_i))| > \delta_3) < \gamma. \qquad (2)$$

Finally we have the bound,

$$\mathbb{P}(\sup_{x_i, x_j \in \mathcal{X}_{n_0}} \left| -\widehat{t}\log(\mathbb{P}(X^n_{g_n^{-2}\widehat{t}} = x_j | X^n_0 = x_i)) - \widehat{t}\log(np(x_j)) - \rho_{\overline{\sigma}(x)}(x_i, x_j)^2 \right| > \delta_1 + \widehat{t}\delta_3) < \gamma$$

To combine the bounds, given some $\delta$ and $\gamma$, set $b_j^n = \log(np(x_j))$, pick $\widehat{t}$ such that $\delta_1 < \delta/2$, then pick $n$ such that the bound in Eq. 2 holds with probability $\gamma$ and error $\delta_3 < \delta/(2\widehat{t})$. □

## 4 Heat kernel for topic models

**Theorem S4.1** (Heat kernel estimates for the topic model). *Let $h$ have smooth subgaussian tails as defined by the following conditions; there exists some $\psi(x)$ such that $\sup_x \psi(x) < \infty$ and $\int_{\mathbb{R}^d} ||x||_2^{2d+4} \psi(x) dx \leq \infty$ such that:*

1. *(tail bound) For all $|\nu| < 6$, $|D_x^\nu h(x)| < \psi(x)$*

2. *(convolved tail bound) For all $|\nu| < 6$, for the $k$-fold convolved kernel $h^{(k)}$, $|D_x^\nu h^{(k)}(x)| < k^\gamma \psi(K^{-\gamma} x)$ for $\gamma > 0$.*

*Further, if $\log(w(x))$ has bounded gradients of order up to 6, then we have the following:*
*Let $\sigma_y^2 = \sigma_0^2 \sigma^2$, where $\sigma_0 = \int ||x||_2^2 h(||x||_2^2) dx$ then the random walk $Y_t$ defined by Equation 4 admits a heat kernel approximation of the marginal distribution at time $t$ in terms of constants $\pi_1(x,y)$ and $v_1(x,y)$,*

$$\sup_{x,y} \left| \mathbb{P}(Y_t = y_t | Y_0 = y_0) - \frac{1}{(4\pi t \sigma_y^2)^{d/2}} \exp\left(\frac{-||y_t - y_0||_2^2}{4\sigma_y^2 t}\right) \left(1 + v_1(y_t, y_0)\sigma_y^2 t + o(\sigma_y^2 t)\right) \right| \leq t^{-1/2} \pi_1(y_t, y_0) + o(t^{-1})$$

*Proof.* First, the Stroock-Varadhan theorem [17] implies that after $t = \widehat{t}\sigma^{-2}$ steps there exists a limiting process $\lim_{\sigma \to 0} Y_{\widehat{t}\sigma^{-2}} \to \widehat{Y}_{\widehat{t}}$ described by the SDE

$$d\widehat{Y}_{\widehat{t}} = \nabla \log(w(\widehat{Y}_{\widehat{t}}))\sigma_0^2 d\widehat{t} + \sigma_0 dW_{\widehat{t}}.$$

In our case, we can determine the rate of convergence of the marginal distributions of $Y_t$ to $\widehat{Y}_t$ by an Edgeworth approximation due to our smooth tail constraints on $h$[5, Theorem 4.1].

$$\sup_{y_t, y_0} \left| \mathbb{P}(Y_t = y_t | Y_0 = y_0) - \mathbb{P}(\widehat{Y}_t = y_t | \widehat{Y}_0 = y_0) - t^{-1/2}\pi_1(y_t, y_0) - t^{-1}\pi_2(y_t, y_0) \right| \leq O(t^{-1-\delta})$$

The details of $\pi_1(y_t, y_0)$ and $\pi_2(y_t, y_0)$ are given in [5, Theorem 4.1], we note that if the drift is constant $\nabla \log(w(x)) = c$, the marginal of $\widehat{Y}_t$ is exactly gaussian and $\pi_1(y_t, y_0)$ and $\pi_2(y_t, y_0)$ are exactly the terms in an Edgeworth approximation when applying the central limit theorem to $h(x)$.

The above approximation is tight as $t \to \infty$; however, the marginal distribution of $\widehat{Y}_{t\sigma^2}$ is only Gaussian as $t\sigma^2 \to 0$. We show that this convergence is fast in $t\sigma^2$ such that if $\sigma^2$ is sufficiently small the heat kernel is still an useful approximation.

Let $\widehat{q}_t(y, x) = \mathbb{P}(\widehat{Y}_t = y | \widehat{Y}_0 = x)$ then by the Fokker-Planck equation, this fulfils the following relationship:

$$\frac{\partial}{\partial t}\widehat{q}_t(y_t, y_0) = \sum_{i,j} \frac{\partial}{\partial y_j \partial y_i}\sigma_0^2 \widehat{q}_t(y_t, y_0) + \sum_i \frac{\partial}{\partial y_i}\nabla \log(w(y_t))\sigma_0^2 \widehat{q}_t(y_t, y_0)$$

We use short time asymptotics of second-order elliptic differential equations to obtain the higher order expansion [4]:

$$\widehat{q}(\widehat{t}, x, y) = \frac{1}{(4\pi \sigma_0^2 \widehat{t})^{d/2}} \exp\left(-\frac{||x_i - y_i||_2^2}{4\sigma_0^2 \widehat{t}}\right) \left(\sum_{j=0}^{\infty} v_j(x, y) t^j\right)$$

Recall that $\widehat{t} = \sigma^2 t$. Substituting into the above gives that

$$\sup_{x,y} \left| \mathbb{P}(Y_t = y_t | Y_0 = y_0) - \frac{1}{(4\pi t \sigma_y^2)^{d/2}} \exp\left(\frac{-||y_t - y_0||_2^2}{4\sigma_y^2 t}\right) \left(\sum_{j=0}^{\infty} v_j(y_t, y_0) t^j\right) \right| \leq t^{-1/2}\pi_1(y_t, y_0) + o(t^{-1})$$

Finally it suffices to show that $v_0(x,y) = 1$ which follows from the fact that our data lie in Euclidean space[16]. In more general manifolds, there will be some curvature associated distortion to the density. □

This proof gives the intuition behind Varadhan's formula. While there are confounders such as the kernel $h$, drift $w$, and curvature $\text{Hess}(\log w)$; these issues all dissapear when $t \to \infty$ (large window size) and $\sigma^{-2} << t$ (topics remain local).

Combining this with the emission probability of $X$ gives the appropriate heat kernel estimate directly on the observed random walk over words. Applying Theorem S4.1, we obtain the following approximation:

$$\mathbb{P}(X_t = x_j | X_0 = x_i) = \int \int \mathbb{P}(X_t = x_j | Y_t = y_t) \mathbb{P}(Y_t = y_t | Y_0 = y_0) \mathbb{P}(Y_0 = y_0 | X_0 = x_i) dy_0 dy_t$$

$$= \frac{1}{\pi_X(x_i)} \int \int \mathbb{P}(X_t = x_j | Y_t = y_t) \mathbb{P}(X_i = x_i, Y_0 = y_0) \mathbb{P}(Y_t = y_t | Y_0 = y_0) dy_0 dy_t$$

Where $\pi_X(x_i)$ is the unigram frequency. Dealing with the inner integral first,

$$\int \mathbb{P}(X_i = x_i, Y_0 = y_0) \mathbb{P}(Y_t = y_t | Y_0 = y_0) dy_0$$

$$\propto \alpha_i \int w(y_0) \exp\left(-\frac{||x_i - y_0||_2^2}{\overline{\sigma}^2}\right) \frac{1}{(4\pi\sigma_y^2 t)^{d/2}} \exp\left(-\frac{||y_t - y_0||_2^2}{4\sigma_y^2 t}\right)(1 + o(t)) dy_0$$

$$= \alpha_i w(x_i)(2\pi\overline{\sigma})^{d/2} \frac{1}{(4\pi\sigma_y^2 t)^{d/2}} \exp\left(-\frac{||y_t - x_i||_2^2}{4\sigma_y^2 t}\right)(1 + O(\sigma_y t) + O(\overline{\sigma}^2) + O(t^{-1/2}))$$

Where the last approximation is a Laplace approximation for small $\overline{\sigma}$ taken at $x_i$ [3]. Now applying the integral over $y_t$

$$\mathbb{P}(X_t = x_j | X_0 = x_i) \propto \frac{\alpha_i}{\pi(x_i)} w(x_i) \exp\left(-\frac{||x_j - x_i||_2^2}{2(\overline{\sigma}^2 + t\sigma_y^2)}\right)(1 + O(\sigma_y^2 t) + O(\overline{\sigma}^2)) + O(t^{-1/2})$$

This has the appropriate form of a heat kernel estimate with the $a_i = \log(\alpha_i) + \log(w(x_i)) - \log(\pi(x_i))$ with two sources of error: too few steps resulting in non-gaussian transitions $O(t^{-1/2})$ and too many steps introducing distortions $O(\sigma_y^2 t)$, $O(\overline{\sigma}^2)$.

## 4.1 Relationship to the topic model of Arora et al

The preprint [1] suggests a latent topic model and consider the following model. Define $c_t$ a discrete-time continuous space latent topic process with the following restrictions:

1. (Stationary distribution near zero) the stationary distribution $C$ is a product distribution, and

$$\mathbb{E}_{c \sim C}[|c_i|^2] = 1/d \qquad \text{and almost surely} \qquad |c_i| \leq 2/\sqrt{d}$$

2. (Increments of $c$ have light tails and converge to zero for large corpora)

$$\mathbb{E}_{p(c_{t+1}|c_t)}[\exp(4\kappa|c_{t+1} - c_t|_1 \log(m))] \leq 1 + \varepsilon_2$$

The observed sentence is then generated by

$$P(w|c) = \frac{\exp(\langle v_w, c \rangle)}{\sum_w \exp(\langle v_w, c \rangle)}$$

Under these conditions, they show that for words $w, w'$ and for sufficiently large corpora,

$$\log(p(w, w') = \frac{1}{2d}||v_w + v'_w||^2 - 2\log Z \pm o(1)$$

This model is qualitatively quite similar to our topic model. Condition (1) on the stationary distribution is analogous to our limit $\bar{\sigma} \to 0$, which ensures the noise term is sharp with respect to our stationary distribution $w(x)$. Condition (2) is the increment size constraint, $g_n \to 0$.

The conceptual distinction between these two methods is that our topic model arises as a natural extension of our short time asymptotic manifold analysis. The heat kernel argument gives direct intuition and justification for the Gaussian decay of the resulting marginal distribution.

Examining the models in detail, the two conditions of [1] on the latent topic model are stronger than ours in the sense that we do not require quantitative bounds on the stationary distribution or the increment size; they may go to zero at any rate with respect to the corpus size. We gain these weaker conditions by assuming that the vocabulary size ($n$ in our notation) goes to infinity and taking many steps $t \to \infty$.

This trade-off between additional assumptions either as direct constraints or additional limits is unavoidable. Recall that

$$\mathbb{P}(X_t = x_j | X_0 = x_i) = \frac{1}{\pi_X(x_i)} \int \int w(y_0) \mathbb{P}(X_t = x_j | Y_t = y_t) \mathbb{P}(X_i = x_i | Y_0 = y_0) \mathbb{P}(Y_t = y_t | Y_0 = y_0) dy_0 dy_t.$$

In order to obtain exponential decay on the LHS assuming only exponential decay in the word emissions $\mathbb{P}(X_i = x_i | Y_0 = y_0)$, we must either invoke a Guassian limit for $\mathbb{P}(Y_t = y_t | Y_0 = y_0)$ or converge it to a point mass relative to $\mathbb{P}(X_i = x_i | Y_0 = y_0)$. Our use of the large vocabulary limit and the heat-kernel approximation allows us to take the former limit, rather than use assumptions to force $\mathbb{P}(Y_t = y_t | Y_0 = y_0)$ to a point mass.

# 5 Empirical evaluation details

## 5.1 Implementation details

We used off-the-shelf available implementations of `word2vec`[*] and `GloVe`[†]. In the paper, these two methods are run with their standard settings, with two exceptions: GloVe's corpus weighting is disabled, as this generally produced superior results, and GloVe's stepsizes are reduced as the default stepsized resulted in NaN-valued embeddings.

For all the models we used 300-dimensional vectors, with window size 5. For `word2vec` we used the skip-gram version with 5 negative samples, 10 iterations, $\alpha = 0.025$ and frequent word sub-sampling with a parameter of $10^{-3}$. For `GloVe` we used $X_{\text{MAX}} = 10$, $\eta = 0.01$ and 10 iterations.

The two other methods (randomized) SVD and regression embedding are both implemented on top of the GloVe codebase. For SVD we factorize the PPMI with no shift ($\tau = 0$ in our notation from the main text) using 50,000 vectors in the randomized projection approximation. For regression, we use $\theta = 50$ and $\eta$ is line-searched starting at $\eta = 10$.

### 5.1.1 Regression embedding

For regression embedding, we do standard stochastic gradient descent with two differences: first, any word co-occurence pairs $C_{ij}$ with counts fewer than ten are skipped with probability proportional to $1 - C_{ij}/10$, this is done to achieve dramatic speedups in training time with no detectable loss in accuracy. Second, we avoid the problem of stepsize tuning by using an initial line search step comblateined with a linear stepsize decay by epoch. Otherwise, initialization and other optimizer choices are kept identical to GloVe.

### 5.1.2 Randomized SVD

Due to the memory and runtime requirements of running a full SVD decomposition, we performed approximate SVDs using randomized projections.

For the SVD algorithm of [7], we use the GloVe co-occurence counter combined with a parallel randomized projection based SVD factorizer based upon the redsvd library [‡]. We implement resonable best practices of [8] of using the square root factorization and no negative shifts. For the number of approximation vectors, we tried various sizes and found vector counts past 50,000 offered little improvement.

## 5.2 Word embedding corpora

We used three corpora to train the word embeddings: the full Wikipedia dump of 03/2015 (about 2.4B tokens), a larger corpus similar to that used by GloVe [14]: Wikipedia2015 + Gigaword5 (5.8B tokens in total) and the one used word2vec [10], which consists of a mixture of several corpora from different sources (6.4B tokens in total).

We preprocessed all the corpora by removing punctuation, numbers and lower-casing all the text. Finally we ran two passes of word2vec's tokenizer word2phrase. As a final step, we removed function words from the vocabulary and kept only the 100K most common words for all our experiments.

## 5.3 Datasets for semantic tasks

Our first set of experiments is on two standard open-vocabulary analogy tasks: Google [9] and MSR [11]. Google consists of 19,544 semantic and syntactic analogy questions, while MSR's 8,000 questions are all syntactic. As an additional analogy task, we use the SAT analogy questions (version 3) of Turney [18]. The dataset contains 374 questions from actual SAT exams, guidebooks, from the ETS web site and other sources. Each question consists of 5 exemplar pairs of words *word1:word2*, where all the pairs hold the same relation. The task is to pick from among another five pairs of words the one that best represents the relation represented by the exemplars. To the best of our knowledge, this is the first time word embeddings are used to solve this task.

---

[*]http://code.google.com/p/word2vec
[†]http://nlp.stanford.edu/projects/glove
[‡]https://github.com/ntessore/redsvd-h

Given the current lack of freely available datasets with category and sequence questions, as described in Section 2, we decided to create them. We used `nltk`'s[§] interface to WordNet [12] in combination with word-word PMI values computed on the Wiki corpus to create the sequences and classes.

As a first step, we collected a set of root words from other semantic tasks to initialize the methods. For the classification data, we created the in-category words by selecting words from various WordNet relations associated to the root words, after which we pruned down to four words based on PMI-similarity to the root word and the other words in the class. The additional options for the multiple choice question were created searching over words related to the root by a different relation type, and selecting those most similar to the root.

For the sequence data, we obtained from WordNet trees of words given by various relation types, and then pruned based on similarity to the root word. For the multiple-choice version of the data, we selected additional (incorrect) options by searching over other words related to the root word, and pruning, as for sequences, based on PMI similarity.

Finally, we manually pruned all three sets of questions, keeping only the most coherent questions, in order to increase the quality of the datasets. After pruning, the category dataset was left with 215 questions and the sequence dataset with 51 questions in its open-vocabulary version and 169 in its multiple choice version.

The two datasets will be made available for others to experiment with. We hope that they help broaden the type of tasks used to evaluate semantic content of word embeddings.

## 5.4 Solving classification and series completion tasks

In each task we obtain an ideal point via the following vector operations.

- **Analogies**: Given A:B::C form the ideal point by $B - A + C$ following the parallelogram rule.
- **Analogies (SAT)**: Given A:B and candidates $C_1 : D_1 \ldots C_n : D_n$ form the ideal point by $B - A$ and represent the options as $D_i - C_i$.
- **Categories**: Given a set $w_1, \ldots, w_n$ defining a category, we define the ideal to be $I = \frac{1}{n} \sum_{i=1}^{n} w_i$.
- **Sequence**: Given sequence $w_1 : \cdots : w_n$ we compute the ideal as $I = w_n + \frac{1}{n}(w_n - w_1)$.

## 5.5 Similarity metrics for verbal reasoning task

Given the ideal point $I$ of a task and options (possibly the entire vocabulary) we pick the answer by proximity of the ideal point $I$ measured in three possible ways.

- **Cosine**: We first unit-normalize each vector as $w_i / ||w_i||_2$ and use cosine similarity to choose which vector is closest to the ideal.
- **$L_2$**: We do not apply any normalization, and pick the closest vector by $L_2$ distance.
- **Diff-cosine (SAT only)**: For the SAT, the differences of the vectors are normalized, and similarity is masured by cosine distance.

In our experiments we found cosine and $L_2$ to give reasonable performance under all tasks. The pre-normalization of cosine vectors are consistent to what was done in [10, 6]). For the $L_2$ distance we applied no normalization.

---

[§]http://www.nltk.org/

# 6 MNIST figure

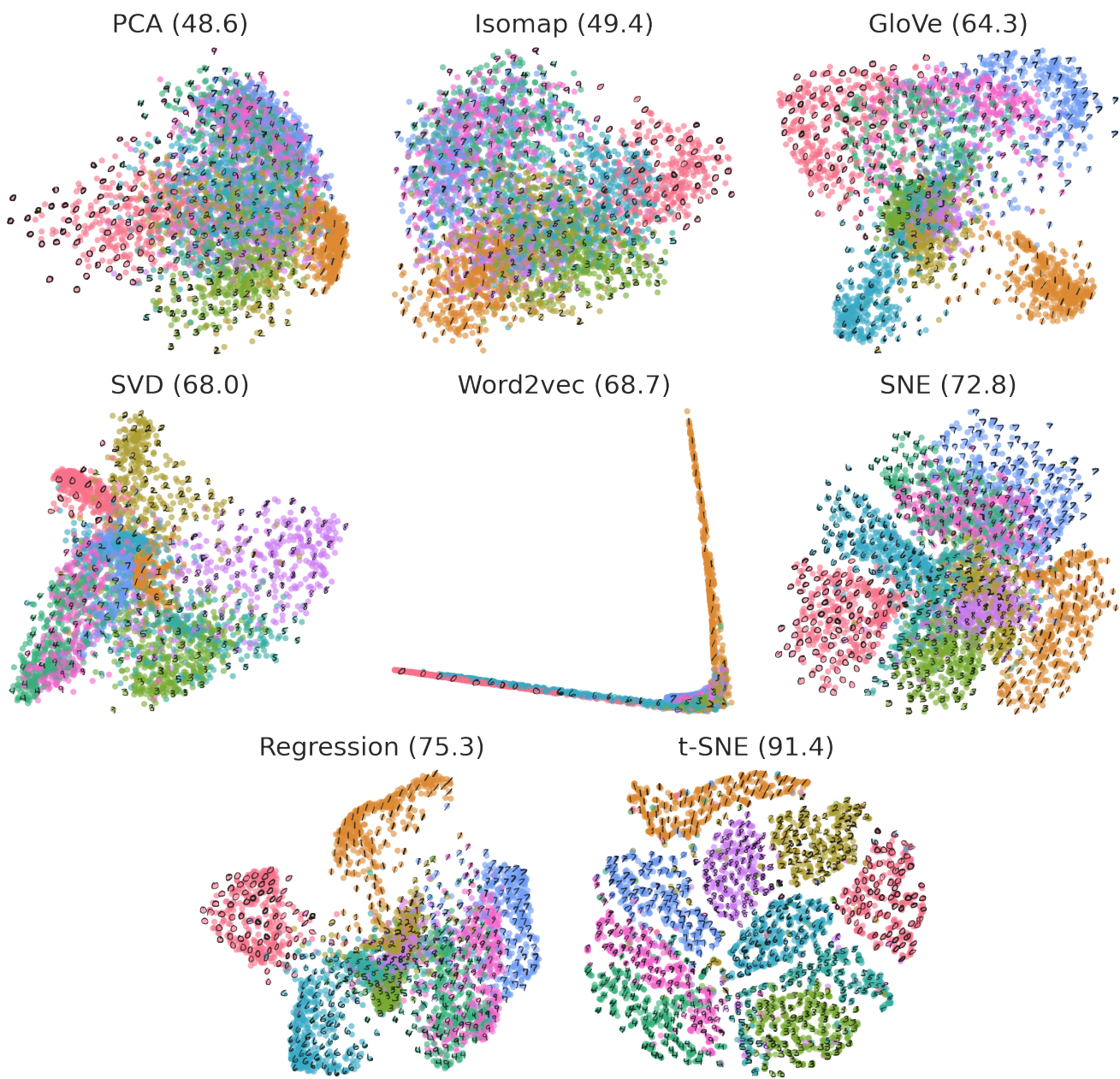

Figure 1: MNIST digit embedding using word embedding method and metric embedding on the same graph.

# 7 Full table of analogy results

## 7.1 Top-30k vocabulary resctriction

|  | Google Analogies | | | SAT | MSR Analogies |
|---|---|---|---|---|---|
|  | Semantic | Syntactic | Total | | |
| Covered | 5022 | 8195 | 13217 | 217 | 4358 |
| Total | 8869 | 10675 | 19544 | 374 | 8000 |

Table 1: glove corpus question coverage

|  | Google Analogies | | | SAT | MSR Analogies |
|---|---|---|---|---|---|
|  | Semantic | Syntactic | Total | | |
| Covered | 4746 | 7679 | 12425 | 199 | 4340 |
| Total | 8869 | 10675 | 19544 | 374 | 8000 |

Table 2: wiki corpus question coverage

|  | Google Analogies | | | SAT | MSR Analogies |
|---|---|---|---|---|---|
|  | Semantic | Syntactic | Total | | |
| Covered | 3965 | 8447 | 12412 | 257 | 4554 |
| Total | 8869 | 10675 | 19544 | 374 | 8000 |

Table 3: w2v corpus question coverage

| Method | Google Analogies (cosine) | | | Google Analogies (l2) | | | SAT | | | MSR Analogies | |
|---|---|---|---|---|---|---|---|---|---|---|---|
|  | Semantic | Syntactic | Total | Semantic | Syntactic | Total | L2 | diff-cosine | cosine | cosine | $L_2$ |
| regression | **78.4** | 70.5 | **73.5** | **74.1** | 70.0 | **71.2** | 38.7 | 41.7 | 33.7 | **67.2** | **64.0** |
| GloVE | 70.2 | 70.9 | 70.6 | 59.2 | 67.7 | 64.5 | 37.2 | 40.7 | 35.7 | 61.2 | 53.5 |
| SVD | 55.8 | 46.4 | 50.0 | 49.1 | 41.2 | 44.3 | 32.7 | 32.2 | 28.1 | 33.5 | 30.3 |
| word2vec | 68.0 | **73.8** | 71.6 | 66.6 | **71.2** | 69.4 | **41.9** | **42.9** | **41.4** | 65.0 | 63.4 |

Table 4: wiki corpus analogy accuracy

| Method | Classification | | Sequence | | Sequence (open vocab) | | Sequence (open vocab, top5) | |
|---|---|---|---|---|---|---|---|---|
|  | Cosine | $L_2$ | Cosine | $L_2$ | Cosine | $L_2$ | Cosine | $L_2$ |
| regression | **86.1** | **85.6** | 58.0 | 55.6 | **7.8** | **5.9** | **72.5** | **60.8** |
| GloVE | 80.9 | 76.7 | **59.2** | 51.5 | 2.0 | 2.0 | 51.0 | 37.3 |
| SVD | 74.9 | 64.7 | 46.2 | 46.2 | 2.0 | 2.0 | 21.6 | 25.5 |
| word2vec | 85.1 | 71.6 | 57.4 | **59.2** | 2.0 | **5.9** | 49.0 | 51.0 |

Table 5: wiki corpus for classification and sequence

| Method | Google Analogies (cosine) | | | Google Analogies (l2) | | | SAT | | | MSR Analogies | |
|---|---|---|---|---|---|---|---|---|---|---|---|
|  | Semantic | Syntactic | Total | Semantic | Syntactic | Total | L2 | diff-cosine | cosine | cosine | $L_2$ |
| regression | **78.4** | 70.8 | **73.7** | **75.5** | 70.9 | **72.6** | 39.2 | 40.6 | 37.8 | 65.6 | 63.9 |
| GloVE | 72.6 | 71.2 | 71.7 | 65.6 | 66.6 | 67.2 | 36.9 | 42.8 | 33.6 | 62.0 | 55.6 |
| SVD | 57.4 | 50.8 | 53.4 | 53.7 | 48.2 | 50.3 | 27.1 | 32.2 | 25.8 | 32.0 | 30.6 |
| word2vec | 73.4 | **73.3** | 73.3 | 71.4 | **70.9** | 71.1 | **42.0** | **49.2** | **42.0** | **67.9** | **66.5** |

Table 6: glove corpus analogy accuracy

|         | Classification |       | Sequence |       | Sequence (open vocab) |       | Sequence (open vocab, top5) |       |
|---------|:------:|:------:|:------:|:------:|:------:|:------:|:------:|:------:|
| Method  | Cosine | $L_2$ | Cosine | $L_2$ | Cosine | $L_2$ | Cosine | $L_2$ |
| regression | **84.6** | **87.6** | **58.9** | **58.3** | 0.0 | 0.0 | 23.5 | 21.6 |
| GloVE      | 80.1 | 73.1 | **58.9** | 48.8 | 0.0 | 0.0 | 27.5 | 23.5 |
| SVD        | 74.6 | 65.2 | 53.0 | 52.4 | 0.0 | 2.0 | 19.6 | 15.7 |
| word2vec   | **84.6** | 76.4 | 56.2 | 54.4 | 0.0 | **3.9** | **53.0** | **58.8** |

Table 7: glove corpus for classification and sequence

|         | Google Analogies (cosine) | | | Google Analogies (l2) | | | SAT | | | MSR Analogies | |
|---------|:------:|:------:|:------:|:------:|:------:|:------:|:------:|:------:|:------:|:------:|:------:|
| Method  | Semantic | Syntactic | Total | Semantic | Syntactic | Total | L2 | diff-cosine | cosine | cosine | $L_2$ |
| regression | **80.1** | 73.0 | **75.2** | **77.3** | **73.1** | **74.4** | 38.1 | 43.0 | 36.9 | 69.4 | 68.4 |
| GloVE      | 70.4 | 73.0 | 72.2 | 61.9 | 70.0 | 67.2 | 36.9 | 43.9 | 34.0 | 66.4 | 61.6 |
| SVD        | 55.2 | 43.6 | 54.1 | 52.8 | 50.6 | 51.3 | 27.9 | 37.3 | 29.1 | 35.6 | 35.4 |
| word2vec   | 66.8 | **73.4** | 71.3 | 67.2 | 72.2 | 70.6 | **39.0** | **46.4** | **42.3** | **75.3** | **75.6** |

Table 8: w2v corpus analogy accuracy

|         | Classification |       | Sequence |       | Sequence (open vocab) |       | Sequence (open vocab, top5) |       |
|---------|:------:|:------:|:------:|:------:|:------:|:------:|:------:|:------:|
| Method  | Cosine | $L_2$ | Cosine | $L_2$ | Cosine | $L_2$ | Cosine | $L_2$ |
| regression | 81.4 | **85.5** | 57.1 | **55.4** | 0.0 | 0.0 | 25.5 | 21.6 |
| GloVE      | 78.2 | 70.0 | **57.7** | 50.6 | 2.0 | 0.0 | 31.4 | 31.4 |
| SVD        | 74.1 | 61.1 | 47.0 | 48.2 | 0.0 | 0.0 | 35.3 | 21.6 |
| word2vec   | **87.0** | 75.0 | 52.7 | 50.9 | **3.9** | **5.9** | **49.0** | **45.1** |

Table 9: w2v corpus for classification and sequence

## 7.2 Top-100k vocabulary

|         | Google Analogies | | | SAT | MSR Analogies |
|---------|:------:|:------:|:------:|:------:|:------:|
|         | Semantic | Syntactic | Total | | |
| Covered | 7829 | 10411 | 18240 | 217 | 5612 |
| Total   | 8869 | 10675 | 19544 | 374 | 8000 |

Table 10: glove corpus question coverage

|         | Google Analogies | | | SAT | MSR Analogies |
|---------|:------:|:------:|:------:|:------:|:------:|
|         | Semantic | Syntactic | Total | | |
| Covered | 7667 | 10231 | 17898 | 199 | 5186 |
| Total   | 8869 | 10675 | 19544 | 374 | 8000 |

Table 11: wiki corpus question coverage

|         | Google Analogies | | | SAT | MSR Analogies |
|---------|:------:|:------:|:------:|:------:|:------:|
|         | Semantic | Syntactic | Total | | |
| Covered | 7213 | 10405 | 17618 | 244 | 5462 |
| Total   | 8869 | 10675 | 19544 | 374 | 8000 |

Table 12: w2v corpus question coverage

|  | Google Analogies (cosine) | | | Google Analogies (l2) | | | SAT | | | MSR Analogies | |
|---|---|---|---|---|---|---|---|---|---|---|---|
| Method | Semantic | Syntactic | Total | Semantic | Syntactic | Total | L2 | diff-cosine | cosine | cosine | $L_2$ |
| regression | **76.9** | 64.6 | **69.9** | 64.9 | 62.5 | 63.5 | 38.7 | 41.7 | 33.7 | **62.6** | 57.4 |
| GloVE | 69.0 | 66.0 | 67.3 | 53.5 | 62.1 | 58.4 | 37.2 | 40.7 | 35.7 | 58.6 | 50.2 |
| SVD | 53.8 | 40.2 | 46.1 | 40.2 | 34.2 | 36.8 | 32.7 | 32.1 | 28.1 | 31.3 | 26.7 |
| word2vec | 67.9 | **70.4** | 69.3 | **67.4** | **67.2** | **67.3** | **41.7** | **43.2** | **41.2** | 62.4 | **61.5** |

Table 13: wiki corpus analogy accuracy

|  | Classification | | Sequence | | Sequence (open vocab, top 5) | | Sequence (open vocab) | |
|---|---|---|---|---|---|---|---|---|
| Method | Cosine | $L_2$ | Cosine | $L_2$ | Cosine | $L_2$ | Cosine | $L_2$ |
| regression | **86.0** | **85.6** | 58.0 | **55.6** | **62.7** | 37.3 | **11.8** | **15.7** |
| GloVE | 80.9 | 76.7 | **59.2** | 51.5 | 51.0 | 37.3 | 3.9 | 3.9 |
| SVD | 74.9 | 64.7 | 46.2 | 46.2 | 21.6 | 25.5 | 3.9 | 3.9 |
| word2vec | 85.1 | 71.6 | 45.1 | 43.1 | 43.1 | **45.1** | 3.9 | 11.8 |

Table 14: wiki corpus for classification and sequence

|  | Google Analogies (cosine) | | | Google Analogies (l2) | | | SAT | | | MSR Analogies | |
|---|---|---|---|---|---|---|---|---|---|---|---|
| Method | Semantic | Syntactic | Total | Semantic | Syntactic | Total | L2 | diff-cosine | cosine | cosine | $L_2$ |
| regression | **75.0** | 66.4 | 70.1 | **70.0** | 66.1 | 67.7 | 39.2 | 40.6 | 37.8 | 62.2 | 58.9 |
| GloVE | 70.7 | 67.5 | 68.8 | 62.5 | 62.4 | 62.5 | 36.9 | 42.9 | 33.6 | 61.0 | 53.0 |
| SVD | 57.0 | 44.2 | 50.3 | 47.9 | 42.0 | 44.5 | 27.2 | 32.3 | 25.8 | 30.6 | 27.4 |
| word2vec | 71.7 | **71.5** | **71.5** | **70.0** | **68.7** | **69.5** | **42.1** | **48.2** | **41.7** | **67.0** | **66.8** |

Table 15: glove corpus analogy accuracy

|  | Classification | | Sequence | | Sequence (open vocab top 5) | | Sequence (top 1) | |
|---|---|---|---|---|---|---|---|---|
| Method | Cosine | $L_2$ | Cosine | $L_2$ | Cosine | $L_2$ | Cosine | $L_2$ |
| regression | **84.6** | **87.6** | **58.9** | **58.3** | 23.5 | 17.6 | 0.0 | 0.0 |
| GloVE | 80.1 | 73.1 | 58.3 | 48.8 | 27.5 | 23.5 | 0.0 | 0.0 |
| SVD | 74.6 | 65.1 | 55.6 | 54.4 | 19.6 | 11.8 | 0.0 | 3.9 |
| word2vec | **84.6** | 76.4 | 55.6 | 54.4 | **49.0** | **54.9** | 0.0 | **7.8** |

Table 16: glove corpus for classification and sequence

|  | Google Analogies (cosine) | | | Google Analogies (l2) | | | SAT | | | MSR Analogies | |
|---|---|---|---|---|---|---|---|---|---|---|---|
| Method | Semantic | Syntactic | Total | Semantic | Syntactic | Total | L2 | diff-cosine | cosine | cosine | $L_2$ |
| regression | **78.2** | 68.9 | **72.7** | **72.0** | 68.6 | **70.0** | 38.1 | 43.0 | 36.9 | 66.1 | 63.1 |
| GloVE | 70.6 | **69.8** | 70.1 | 61.2 | 65.7 | 63.9 | 36.9 | **53.9** | 34.0 | 65.3 | 59.0 |
| SVD | 55.9 | 47.8 | 51.1 | 45.4 | 44.7 | 45.0 | 27.9 | 37.2 | 29.1 | 33.6 | 31.1 |
| word2vec | 67.1 | 71.6 | 69.8 | 68.0 | **70.4** | 69.4 | **39.2** | 47.1 | **42.8** | **73.8** | **74.6** |

Table 17: w2v corpus analogy accuracy

|  | Classification | | Sequence | | Sequence (top 5) | | Sequence (top 1) | |
|---|---|---|---|---|---|---|---|---|
| Method | Cosine | $L_2$ | Cosine | $L_2$ | Cosine | $L_2$ | Cosine | $L_2$ |
| regression | 81.3 | **85.5** | 57.1 | **55.4** | 24.5 | 21.6 | 0.0 | 0.0 |
| GloVE | 78.2 | 70.0 | **58.3** | 50.6 | 31.4 | 31.4 | 3.9 | 0.0 |
| SVD | 74.1 | 61.1 | 45.8 | 48.2 | 31.4 | 21.6 | 0.0 | 0.0 |
| word2vec | **87.0** | 75.0 | 53.3 | 50.9 | **43.1** | **35.3** | **7.84** | **11.8** |

Table 18: w2v corpus for classification and sequence